\pdfoutput=1

\documentclass[11pt]{article}

\usepackage{naacl2021}

\usepackage{times}
\usepackage{latexsym}

\usepackage{amsmath,amsfonts,bm}

\def\eqref#1{equation~\ref{#1}}

\def\1{\bm{1}}

\def\vu{{\bm{u}}}

\def\mX{{\bm{X}}}

\DeclareMathAlphabet{\mathsfit}{\encodingdefault}{\sfdefault}{m}{sl}
\SetMathAlphabet{\mathsfit}{bold}{\encodingdefault}{\sfdefault}{bx}{n}

\usepackage[T1]{fontenc}

\usepackage[utf8]{inputenc}

\usepackage{microtype}

\usepackage{hyperref}
\usepackage{url}
\usepackage{graphicx}

\usepackage{multicol}
\usepackage{tabularx}
\usepackage{booktabs}
\usepackage{adjustbox}
\usepackage{url}
\usepackage{xcolor}
\usepackage{amsmath}

\newcommand\alexa{$^\diamondsuit$}
\newcommand\cmu{$^\spadesuit$}
\newcommand\aspace{\hspace{1.05em}}

\title{Example-Driven Intent Prediction with Observers}

\author{Shikib Mehri\cmu\thanks{ Work done while Shikib was at Amazon}\aspace Mihail Eric\alexa  \\
\hspace{2em} \cmu Language Technologies Institute, Carnegie Mellon University, \alexa Amazon Alexa AI \\
\hspace{4em}\texttt{amehri@cs.cmu.edu, mihaeric@amazon.com}\\ 
}

\begin{document}

\maketitle

\begin{abstract}
A key challenge of dialog systems research is to effectively and efficiently adapt to new domains. A scalable paradigm for adaptation necessitates the development of generalizable models that perform well in few-shot settings. In this paper, we focus on the intent classification problem which aims to identify user intents given utterances addressed to the dialog system. We propose two approaches for improving the generalizability of utterance classification models: (1) observers and (2) example-driven training. Prior work has shown that BERT-like models tend to attribute a significant amount of attention to the [CLS] token, which we hypothesize results in diluted representations. Observers are tokens that are not attended to, and are an alternative to the [CLS] token as a semantic representation of utterances. Example-driven training learns to  classify utterances by comparing to examples, thereby using the underlying encoder as a sentence similarity model. These methods are complementary; improving the representation through observers allows the example-driven model to better measure sentence similarities. When combined, the proposed methods attain state-of-the-art results on three intent prediction datasets (\textsc{banking77}, \textsc{clinc150}, \textsc{hwu64}) in both the full data and few-shot (10 examples per intent) settings. Furthermore, we demonstrate that the proposed approach can transfer to new intents and across datasets without any additional training.

\end{abstract}

\section{Introduction}


Task-oriented dialog systems aim to satisfy a user goal in the context of a specific task such as booking flights \citep{hemphill-etal-1990-atis}, providing transit information \citep{raux2005let}, or acting as a tour guide \citep{budzianowski2018multiwoz}. Task-oriented dialog systems must first \textit{understand} the user's goal by extracting meaning from a natural language utterance. This problem is known as \textit{intent prediction} and is a vital component of task-oriented dialog systems \citep{hemphill-etal-1990-atis,coucke2018snips}. Given the vast space of potential domains, a key challenge of dialog systems research is to effectively and efficiently adapt to new domains \citep{rastogi2019towards}. Rather than adapting to new domains by relying on large amounts of domain-specific data, a scalable paradigm for adaptation necessitates the development of generalizable models that perform well in few-shot settings \citep{casanueva2020efficient,dialoglue}.

The task of intent prediction can be characterized as a two step process: (1) \textbf{representation} (mapping a natural language utterance to a semantically meaningful representation) and (2) \textbf{prediction} (inferring an intent given a latent representation). These two steps are complementary and interdependent, thereby necessitating that they be jointly improved.  Therefore, to enhance the domain adaptation abilities of intent classification systems we propose to (1) improve the representation step through \textbf{observers} and (2) improve the prediction step through \textbf{example-driven training}. 

While BERT \citep{devlin2018bert} is a strong model for natural language understanding tasks \citep{wang2018glue}, prior work has found a significant amount of BERT's attention is attributed to the \textit{[CLS]} and \textit{[SEP]} tokens, though these special tokens do not attribute much attention to the words of the input until the last layer \citep{clark2019does,kovaleva2019revealing}. Motivated by the concern that attending to these tokens is causing a dilution of representations, we introduce \textbf{observers}. Rather than using the latent representation of the \textit{[CLS]} token, we instead propose to have tokens which \textit{attend to the words of the input} but \textit{are not attended to}. In this manner, we disentangle BERT's attention with the objective of improving the semantic content captured by the utterance representations.

A universal goal of language encoders is that inputs with similar semantic meanings have similar latent representations \citep{devlin2018bert}. To maintain consistency with this goal, we introduce \textbf{example-driven training} wherein an utterance is classified by measuring similarity to a set of examples corresponding to each intent class. While standard approaches implicitly capture the latent space to intent class mapping in the learned weights (i.e., through a classification layer), example-driven training makes the prediction step an explicit non-parametric process that reasons over a set of examples. By maintaining consistency with the universal goal of language encoders and explicitly reasoning over the examples, we demonstrate improved generalizability to unseen intents and domains.

By incorporating both observers and example-driven training on top of the \textsc{ConvBERT} model\footnote{\url{https://github.com/alexa/DialoGLUE/}} \citep{dialoglue}, we attain state-of-the-art results on three intent prediction datasets:  \textsc{banking77} \citep{casanueva2020efficient}, \textsc{clinc150} \citep{larson-etal-2019-evaluation}, and \textsc{hwu64} \citep{liu2019benchmarking} in both full data and few-shot settings. To measure the generalizability of our proposed models, we carry out experiments evaluating their ability to transfer to new intents and across datasets. By simply modifying the set of examples during evaluation and without any additional training, our example-driven approach attains strong results on both transfer to unseen intents and across datasets. This speaks to the generalizability of the approach. Further, to demonstrate that observers mitigate the problem of diluted representations, we carry out probing experiments and show that the representations produced by observers capture more semantic information than the \textit{[CLS]} token.

The contributions of this paper are as follows: (1) we introduce observers in order to avoid the potential dilution of BERT's representations, by disentangling the attention, (2) we introduce example-driven training which explicitly reasons over a set of examples to infer the intent, (3) by combining our proposed approaches, we attain state-of-the-art results across three datasets on both full data and few-shot settings, and (4) we carry out experiments demonstrating that our proposed approach is able to effectively transfer to unseen intents and across datasets without any additional training.

\section{Methods}

In this section, we describe several methods for the task of intent prediction. We begin by describing two baseline models: a standard BERT classifier \citep{devlin2018bert} and \textsc{ConvBERT} with task-adaptive masked language modelling \citep{dialoglue}. The proposed model extends the \textsc{ConvBERT} model of \citet{dialoglue} through observers and example-driven training. Given the aforementioned two step characterization of intent prediction, observers aim to improve the representation step while example-driven training improves the prediction step.

\subsection{BERT Baseline}

Across many tasks in NLP, large-scale pre-training has resulted in significant performance gains \citep{wang2018glue,devlin2018bert,radford2018improving}. To leverage the generalized language understanding capabilities of BERT for the task of intent prediction, we follow the standard fine-tuning paradigm. Specifically, we take an off-the-shelf BERT-base model and perform end-to-end supervised fine-tuning on the task of intent prediction.

\subsection{Conversational BERT with Task-Adaptive MLM}

Despite the strong language understanding capabilities exhibited by pre-trained models, modelling dialog poses challenges due to its intrinsically goal-driven, linguistically diverse, and often informal/noisy nature. To this end, recent work has proposed pre-training on open-domain \textit{conversational data} \citep{henderson2019convert,zhang2019dialogpt}. Furthermore, task-adaptive pre-training wherein a model is trained in a self-supervised manner on a dataset prior to fine-tuning on the same dataset, has been shown to help with domain adaptation \citep{mehri2019pretraining,gururangan2020don,dialoglue}. Our models extend the \textsc{ConvBERT} model of \citet{dialoglue} which (1) pre-trained the BERT-base model on a large open-domain dialog corpus and (2) performed task-adaptive masked language modelling (MLM) as a mechanism for adapting to specific datasets.

\begin{figure*}
    \centering
    \includegraphics[width=0.8\textwidth]{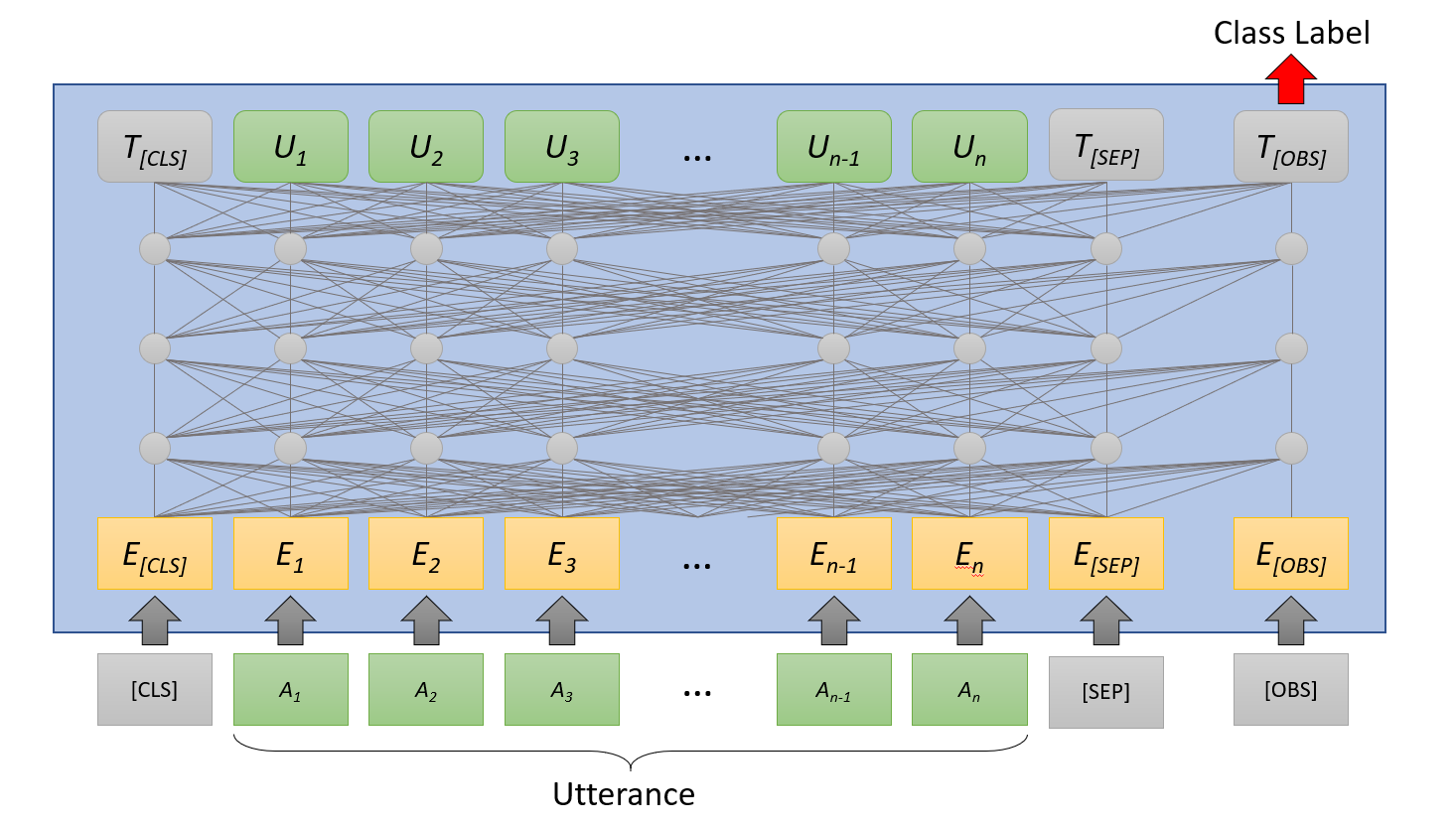}
    \caption{A visualization of the observers. The observer node attends to other tokens at each layer, however it is never attended to. While this figure only depicts one observer -- we include multiple observers and average their final representation.}
    \label{fig:observers}
\end{figure*}

\subsection{Observers}

The pooled representation of BERT-based models is computed using the \textit{[CLS]} token. Analysis of BERT's attention patterns has demonstrated that a significant amount of attention is attributed to the \textit{[CLS]} and \textit{[SEP]} tokens \citep{clark2019does,kovaleva2019revealing}. It is often the case that over half of the total attention is to these tokens \citep{clark2019does}. Furthermore, the \textit{[CLS]} token primarily attends to itself and \textit{[SEP]} until the final layer \citep{kovaleva2019revealing}. It is possible that attending to these special BERT tokens, in combination with the residual connections of the BERT attention heads, is equivalent to a no-op operation. However, it is nonetheless a concern that this behavior of attending to tokens with no inherent meaning (since \textit{[CLS]} does not really attend to other words until the final layer) results in the latent utterance level representations being diluted. 

We posit that a contributing factor of this behavior is the entangled nature of BERT's attention: i.e., the fact that the \textit{[CLS]} token \textit{attends} to words of the input and is \textit{attended to} by the words of the input. This entangled behavior may inadvertently cause the word representations to attend to \textit{[CLS]} in order to better resemble its representation and therefore make it more likely that the \textit{[CLS]} token attends to the word representations. In an effort to mitigate this problem and ensure the representation contains more of the semantic meaning of the utterance, we introduce an extension to traditional BERT fine-tuning called \emph{observers}.

Observers, pictured in Figure \ref{fig:observers}, \textit{attend to the tokens of the input utterance} at every layer of the BERT-based model however they are \textit{never attended to}. The representation of the observers in the last layer is then used as the final utterance level representation. In this manner, we aim to disentangle the relationship between the representation of \textit{each word} in the input and the \textit{final utterance level representation}. By removing this bi-directional relationship, we hope to avoid the risk of diluting the representations (by inadvertently forcing them to attend to a meaningless \textit{[CLS]} token) and therefore capture more semantic information in the final utterance level representation. Throughout our experiments we use $20$ observer tokens (which are differentiated only by their position embeddings) and average their final representations. The positions of the observer tokens is consistent across all utterances (last 20 tokens in the padded sequence). Specifically, the concept of observers modifies $\mathcal{F}$ in Equations 1 and 2. While we maintain the BERT-based model architecture, we instead produce the utterance level representation by averaging the representations of the observer tokens and using that for classification rather than the \emph{[CLS]} token.

\begin{figure*}
    \centering
    \includegraphics[width=0.7\textwidth]{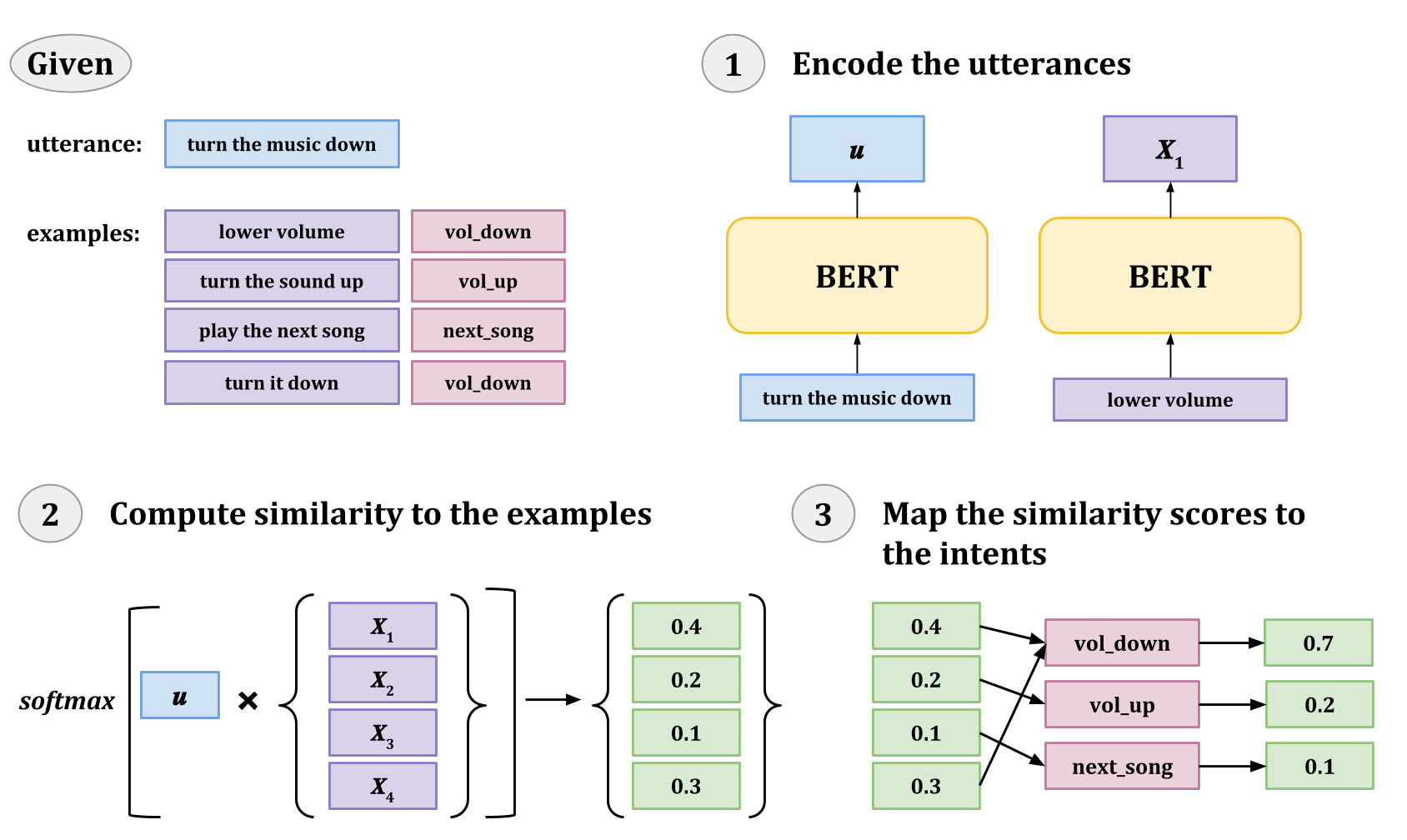}
    \caption{A visualization of the three step process of computing a probability distribution over the set of intents in our example-driven formulation.}
    \label{fig:example}
\end{figure*}
\subsection{Example-Driven Training}

A universal goal of language encoders is for inputs with similar semantic meanings to have similar latent representations. BERT \citep{devlin2018bert} has been shown to effectively identify similar sentences \citep{reimers2019sentence} even without additional fine-tuning \citep{zhang2019bertscore}. Through example-driven training, we aim to reformulate the task of intent prediction to be more consistent with this universal goal of language encoders. 

Using a BERT-like encoder, we train an intent classification model to (1) measure the similarity of an utterance to a set of examples and (2) infer the intent of the utterance based on the similarity to the examples corresponding to each intent. Rather than implicitly capturing the \textit{latent space} to \textit{intent class} mapping in our learned weights (i.e., through a classification layer), we make this mapping an explicit non-parametric process that reasons over a set of examples. Our formulation, similar to metric-based meta learning \citep{koch2015siamese}, only performs gradient updates for the language encoder, which is trained for the task of \textit{sentence similarity}. Through this example-formulation, we hypothesize that the model will better generalize in few-shot scenarios, as well as to rare intents.



We are given (1) a language encoder $\mathcal{F}$ that encodes an utterance to produce a latent representation, (2) a natural language utterance $utt$, and (3) a set of $n$ examples $\{ (\text{x}_1, y_1), \dots, (\text{x}_n, y_n) \}$ where $\text{x}_{1,\dots,n}$ are utterances and $y_{1,\dots,n}$ are their corresponding intent labels. With $\mathcal{F}$ being a BERT-like model, the following equations describe example-driven intent classification:

\begin{gather}
    \vu = \mathcal{F}( utt ) \\
    \mX_i = \mathcal{F}( \text{x}_i ) \\
    \bm{\alpha} = \text{softmax} (\bm{u}^T \cdot \bm{X})\\
    P(c) = \sum_{i:\,y_i = c} \bm{\alpha}_i
\end{gather}

The equations above describe a non-parametric process for intent prediction. Instead, through the example-driven formulation (visualized in Figure \ref{fig:example}), the underlying language encoder (e.g., BERT) is being trained for the task of sentence similarity. A universal goal of language encoders is that inputs with similar semantic meaning should have similar latent representations. By formulating intent prediction as a sentence similarity task, we are adapting BERT-based encoders in a way that is consistent with this universal goal. We hypothesize that in contrast to the baseline models, this formulation facilitates generalizability and has the potential to better transfer to new intents and domains.

At training time, we populate the set of examples in a two step process: (i) for each intent class that exists in the training batch, we sample one \textit{different} utterance of the same intent class from the \textit{ training set} and (ii) we randomly sample utterances from the training set until we have a set of examples that is double the size of the training batch size (128 example utterances). During inference, our example set is comprised of all the utterances in the training data. 

\section{Experiments}

\subsection{Datasets}

We evaluate our methods on three intent prediction datasets: \textsc{banking77} \citep{casanueva2020efficient}, \textsc{clinc150} \citep{larson-etal-2019-evaluation}, and \textsc{hwu64} \citep{liu2019benchmarking}. These datasets span several domains and consist of many different intents, making them more challenging and more reflective of commercial settings than commonly used intent prediction datasets like SNIPs \citep{coucke2018snips}. \textsc{banking77} contains 13,083 utterances related to banking with 77 different fine-grained intents. \textsc{clinc150} contains 23,700 utterances spanning 10 domains (e.g., travel, kitchen/dining, utility, small talk, etc.) and 150 different intent classes. \textsc{hwu64} includes 25,716 utterances for 64 intents spanning 21 domains (e.g., alarm, music, IoT, news, etc.).

\citet{casanueva2020efficient} forego a validation set for these datasets and instead only use a training and testing set. We instead follow the setup of \citet{dialoglue}, wherein a portion of the training set is designated as the validation set. 

\subsection{Experimental Setup}

We evaluate in two experimental settings following prior work \citep{casanueva2020efficient,dialoglue}: (1) using the full training set and (2) using 10 examples per intent or approximately 10\% of the training data. In both settings, we evaluate on the validation set at the end of each epoch and perform early stopping with a patience of 20 epochs for a maximum of 100 epochs. Since the few-shot experiments are more sensitive to initialization and hyperparameters, we repeat the few-shot experiments 5 times and take an average over the experimental runs. For the few-shot settings, our models use \textit{only} the few-shot training data for both masked language modelling and as examples at inference time in the example-driven models (i.e., they do not see any additional data). Our experiments with observers all use 20 observers, however we include an ablation in the appendix (Table \ref{ablation}; see supplementary materials).

\subsection{Results}

Our experimental results, as well as the results obtained by \citet{casanueva2020efficient} and \citet{dialoglue} are shown in Table \ref{results-table}. Combining observers and example-driven training results in (1) SoTA results across the three datasets and (2) a significant improvement over the BERT-base model, especially in the few-shot setting (\textbf{+5.02\%} on average).

Furthermore, the results show that the use of observers is particularly conducive to the example-driven training setup. Combining these two approaches gains strong improvements over the ConvBERT + MLM model (few-shot: \textbf{+4.98\%}, full data: \textbf{+0.41\%}). However, when we consider the two proposed approaches independently, there is no consistent improvement for both example-driven (few-shot: \textbf{-0.46\%} full data: \textbf{+0.24\%}) and observers (few-shot: \textbf{+0\%}, full data: \textbf{-0.42\%}). The fact that these two methods are particularly conductive to each other signifies the importance of using them jointly. The representation step of intent prediction is tackled by observers, which aim to better capture the semantics of an input by disentangling the attention and therefore avoiding the dilution of the representations. The prediction step, is improved through example-driven training which uses the underlying BERT-based model to predict intents by explicitly reasoning over a set of examples. This characterization highlights the importance of jointly addressing both steps of the process simultaneously. Using observers alone does not lead to significant improvements because the linear classification layer cannot effectively leverage the improved representations. Using example-driven training alone does not lead to significant improvements because the \textit{[CLS]} representations do not capture enough of the underlying utterance semantics. The enhanced semantic representation of observers is necessary for example-driven training: by improving the latent representations of utterances, it is easier to measure similarity in the set of examples.

\begin{table*}[!t]
\centering
\renewcommand*{\arraystretch}{1.1}
{
\begin{tabularx}{\linewidth}{l XX XX XX}
\toprule
  {} & \multicolumn{2}{c}{\bf \textsc{banking77}} & \multicolumn{2}{c}{\bf \textsc{clinc150}} & \multicolumn{2}{c}{\bf \textsc{hwu64}} \\
  \cmidrule(lr){2-3} \cmidrule(lr){4-5} \cmidrule(lr){6-7}
\textbf{Model} & \textbf{Few}  & \textbf{Full} & \textbf{Few} & \textbf{Full} & \textbf{Few}  &  \textbf{Full} \\
 \cmidrule(lr){2-3} \cmidrule(lr){4-5} \cmidrule(lr){6-7}

  \cmidrule(lr){1-7} \multicolumn{1}{l}{\textbf{Prior Work}} \\   \cmidrule(lr){1-7}

\textsc{USE*} \citep{casanueva2020efficient} & 84.23 & 92.81  & 90.85 &  95.06  & 83.75 & 91.25 \\
\textsc{ConveRT*} \citep{casanueva2020efficient} & 83.32 & 93.01  & 92.62 & 97.16  & 82.65 & 91.24 \\
\textsc{USE+ConveRT*} \citep{casanueva2020efficient} & 85.19 &  93.36 & 93.26 &  97.16  & 85.83 & 92.62 \\

\textsc{BERT-base} \citep{dialoglue} & 79.87  &  93.02 & 89.52  &  95.93  & 81.69 & 89.97 \\
\textsc{ConvBERT} \citep{dialoglue} & 83.63 &  92.95 & 92.10  &  97.07  & 83.77  & 90.43\\
\textsc{ConvBERT + MLM} \citep{dialoglue} & 83.99  &  93.44 & 92.75  &  97.11  & 84.52 & 92.38 \\

  \cmidrule(lr){1-7} \multicolumn{1}{l}{\textbf{Proposed Models}} \\   \cmidrule(lr){1-7}
\textsc{ConvBERT + MLM + \textit{Example}} & 84.09 & \textbf{94.06} & 92.35  &  97.11  & 83.44  & 92.47 \\
\textsc{ConvBERT + MLM + \textit{Observers}} & 83.73 &  92.83 & 92.47  &  96.76  &  85.06 & 92.10 \\
\textsc{ConvBERT + MLM + \textit{Example} + \textit{Observers}} & \textbf{85.95} &  93.83 & \textbf{93.97} &  \textbf{97.31}  &  \textbf{86.28} & \textbf{93.03} \\

\bottomrule
\end{tabularx}}%
\caption{Accuracy scores ($\times$100\%) on all three intent detection data sets with varying number of training examples (\textbf{Few:} 10 training utterances per intent; \textbf{Full:} full training data). The full data results of \citet{casanueva2020efficient} are trained on more data as they forego a validation set. We follow the setup of \citet{dialoglue}, wherein a portion of the training set is used as the validation set. Results in bold-face are statistically significant by t-test ($p < 0.01$).}
\label{results-table}
\end{table*}

\section{Analysis}

This section describes several experiments that were carried out to show the unique benefits of observers and example-driven training, as well as to validate our hypothesis regarding the two methods. First, we show that with the example-driven formulation for intent prediction, we can attain strong performance on intents unseen during training. Next, we show that the generalization to new intents transfers across datasets. Next, we carry out a probing experiment that demonstrates that the latent representation of the observers contains greater semantic information about the input. Finally, we discuss an ablation over the number of observers used which demonstrates that the benefit of observers is primarily a consequence of the disentangled attention.

\begin{table*}[!t]
\centering
\renewcommand*{\arraystretch}{1.1}
{
\begin{tabularx}{\linewidth}{l X  X  X}
\toprule
  \textbf{Model} & \multicolumn{1}{c}{\bf \textsc{banking77}} & \multicolumn{1}{c}{\bf \textsc{clinc150}} & \multicolumn{1}{c}{\bf \textsc{hwu64}} \\
  \cmidrule(lr){2-2} \cmidrule(lr){3-3} \cmidrule(lr){4-4}

 &    & &  \\[-2ex] 
 
\textsc{BERT-base (off-the-shelf)}  & \multicolumn{1}{c}{19.50}   & \multicolumn{1}{c}{26.50}  & \multicolumn{1}{c}{26.56} \\
\textsc{ConvBERT (off-the-shelf)} & \multicolumn{1}{c}{19.50}   & \multicolumn{1}{c}{26.50}  & \multicolumn{1}{c}{26.56} \\
\textsc{ConvBERT + MLM + \textit{Example}} &  \multicolumn{1}{c}{67.36}   & \multicolumn{1}{c}{79.69}  & \multicolumn{1}{c}{62.24} \\
\textsc{ConvBERT + MLM + \textit{Example} + \textit{Observers}} & \multicolumn{1}{c}{\textbf{84.87}}   & \multicolumn{1}{c}{\textbf{94.35}}  & \multicolumn{1}{c}{\textbf{85.32}} \\

\cmidrule(lr){1-4}

\textsc{Best Fully Trained Model} & \multicolumn{1}{c}{85.95} &  \multicolumn{1}{c}{93.97} & \multicolumn{1}{c}{86.28} \\

\bottomrule
\end{tabularx}}%
\caption{Accuracy scores ($\times$100\%) for transferring to unseen intents averaged over 30 runs wherein 4-10 intents are removed from the few-shot setting during training and added back in during evaluation. The last row corresponds to the best results that were trained with all of the intents, shown in Table \ref{results-table}. Note that the non example-driven models are incapable of predicting unseen slots, and their perform is equivalent to random chance.}
\label{intent-transfer-table}
\end{table*}
\subsection{Transfer to Unseen Intents}

By formulating intent prediction as a sentence similarity task, the example-driven formulation allows for the potential to predict intents that are unseen at training time. We carry out experiments in the few-shot setting for each dataset, by (1) randomly removing 4 - 10 intent classes when training in an example-driven manner, (2) adding the removed intents back to the set of examples during evaluation and (3) reporting results only on the unseen intents. We repeat this process 30 times for each dataset and the results are reported in Table \ref{intent-transfer-table}. It should be noted that we do not perform MLM training on the utterances corresponding to the unseen intents.

These results demonstrate that the example-driven formulation generalizes to new intents, without having to re-train the model. The performance on the unseen intents approximately matches the performance of the best model which has seen all intents (denoted \textsc{Best Fully trained model} in Table \ref{intent-transfer-table}). These results highlight a valuable property of the proposed formulation: namely, that new intent classes can be added in an online manner without having to re-train the model. While the off-the-shelf BERT-base and \textsc{ConvBERT} models, which are not at all fine-tuned on the datasets, are able to identify similar sentences to some extent -- training in an example-driven manner drastically improves performance.

The addition of observers, in combination with example-driven training, significantly improves performance on this experimental setting (\textbf{+18.42\%}). This suggests that the observers generalize better to unseen intents, potentially because the observers are better able to emphasize words that are key to differentiating between intents (e.g., \textit{turn the volume \textbf{up}} vs \textit{turn the volume \textbf{down}}). 
\subsection{Transfer Across Datasets}

While transferring to unseen intents is a valuable property, the unseen intents in this experimental setting are still from the same domain. To further evaluate the generalizability of our models, we carry out experiments evaluating the ability of models to transfer to \textit{other datasets}. Using the full data setting with 10 training utterances per intent, we (1) train a model on a dataset and (2) evaluate the models on a new dataset, using the training set of the new dataset as examples during inference. In this manner, we evaluate the ability of the models to transfer to unseen intents and domains without additional training. 

\begin{table*}[!t]
\centering
\renewcommand*{\arraystretch}{1.1}
{
\begin{tabularx}{\linewidth}{l X  X  X}
\toprule
  \textbf{Model} & \multicolumn{1}{c}{\bf \textsc{banking77}} & \multicolumn{1}{c}{\bf \textsc{clinc150}} & \multicolumn{1}{c}{\bf \textsc{hwu64}} \\
  \cmidrule(lr){2-2} \cmidrule(lr){3-3} \cmidrule(lr){4-4}

 &    & &  \\[-2ex] 
 
\textsc{Trained on banking77}  & \multicolumn{1}{c}{\textit{93.83}}   & \multicolumn{1}{c}{91.26}  & \multicolumn{1}{c}{83.64} \\
\textsc{Trained on clinc150} & \multicolumn{1}{c}{\textbf{85.84}}   & \multicolumn{1}{c}{\textit{97.31}}  & \multicolumn{1}{c}{\textbf{86.25}} \\
\textsc{Trained on hwu64} &  \multicolumn{1}{c}{77.95}   & \multicolumn{1}{c}{\textbf{92.47}}  & \multicolumn{1}{c}{\textit{93.03}} \\

\bottomrule
\end{tabularx}}%
\caption{Accuracy scores ($\times$100\%) for transferring across datasets (in the full data setting) using the ConvBERT + MLM + Example + Observers model. The diagonal consists of results where the model was trained and evaluated on the same dataset.}
\label{dataset-transfer-table}
\end{table*}

\begin{table*}[!t]
\centering
\renewcommand*{\arraystretch}{1.1}
{
\begin{tabularx}{\linewidth}{l X  X  X}
\toprule
  \textbf{Model} & \multicolumn{1}{c}{\bf \textsc{banking77}} & \multicolumn{1}{c}{\bf \textsc{clinc150}} & \multicolumn{1}{c}{\bf \textsc{hwu64}} \\
  \cmidrule(lr){2-2} \cmidrule(lr){3-3} \cmidrule(lr){4-4}

 &    & &  \\[-2ex] 
 
\textsc{ConvBERT + MLM + \textit{Example}} &  \multicolumn{1}{c}{34.22}   & \multicolumn{1}{c}{31.92}  & \multicolumn{1}{c}{19.73} \\
\textsc{ConvBERT + MLM + \textit{Example} + \textit{Observers}} & \multicolumn{1}{c}{\textbf{35.34}}   & \multicolumn{1}{c}{\textbf{33.84}}  & \multicolumn{1}{c}{\textbf{21.19}} \\

\bottomrule
\end{tabularx}}%
\caption{Micro-averaged F-1 scores for the task of reproducing the words of the input (using only the most frequent 1000 words) given the different latent representations.}
\label{observers-experiment}
\end{table*}
The results in Table \ref{dataset-transfer-table} demonstrate the ability of the the model with obsevers and example-driven training to transfer to new datasets, which consist of both unseen intents and unseen domains. These results show that the example-driven model performs reasonably well even when transferring to domains and intents that were not seen at training time. These results, in combination with the results shown in Table \ref{intent-transfer-table} speak to the generalizability of the proposed methods. Specifically, by formulating intent prediction as a sentence similarity task through example-driven training, we are maintaining consistency with a universal goal of language encoders (i.e., that utterances with similar semantic meanings have similar latent representations) that effectively transfers to new settings.

\subsection{Observers Probing Experiment}

We hypothesized that by disentangling the attention in BERT-based models, the observers would avoid the dilution of representations (which occurs because words attend to a meaningless \textit{[CLS]} token) and therefore better capture the semantics of the input. We validate this hypothesis through the experimental evidence presented in Table \ref{intent-transfer-table} wherein the use of observers results in a significant performance improvement on unseen intents. To demonstrate that observers better capture the semantics of an input, we carry out a probing experiment using the \textit{word-content} task of \citet{conneau-etal-2018-cram}.

We generate a latent representation of each utterance using models \textbf{with} and \textbf{without} observers. We then train a classifier layer on top of the frozen representations to reproduce the words of the input. Similar to \citet{conneau-etal-2018-cram}, we avoid using the entire vocabulary for this probing experiment and instead use only the most frequent 1000 words for each dataset. With infrequent words, there would be uncertainty about whether the performance difference is a consequence of (1) the semantic content of the representation or (2) the quality of the probing model. Since we are concerned with measuring the former, we only consider the most frequent words to mitigate the effect of latter. Table \ref{observers-experiment} shows the micro-averaged F-1 score for the task of reproducing the words in the utterance, given the different latent representations.

A latent representation that better captures the semantics of the input utterance, will be better able to reproduce the specific words of the utterance. The results in Table \ref{observers-experiment} show that the use of observers results in latent representations that better facilitate the prediction of the input words (\textbf{+1.50} or \textbf{5\%} relative improvement). These results further validate the hypothesis that the use of observers results in better latent representations.

\subsection{Number of Observers}

To further understand the performance of the observers, we carry out an ablation study over the number of observers. The results shown in Table \ref{ablation} (in the Appendix) demonstrate that while multiple observers help, even a single observer provides benefit. This suggests that the observed performance gain is a primarily a consequence of the disentangled attention rather than averaging over multiple observers. This ablation provides further evidence that the use of observers mitigates the dilution of the utterance level representations.

\section{Related Work}

\subsection{Intent Prediction}

Intent prediction is the task of converting a user's natural language utterance into one of several pre-defined classes, in an effort to describe the user's intent \citep{hemphill-etal-1990-atis,coucke2018snips}. Intent prediction is a vital component of pipeline task-oriented dialog systems, since determining the goals of the user is the first step to producing an appropriate response \citep{raux2005let,young2013pomdp}. Prior to the advent of large-scale pre-training \citep{devlin2018bert,radford2018improving}, approaches for intent prediction utilize task-specific architectures and training methodologies (e.g., multi-tasking, regularization strategies) that aim to better capture the semantics of the input \citep{bhargava2013easy,hakkani2016multi,gupta2018efficient,niu2019novel}. 

The large-scale pre-training of BERT makes it more effective for many tasks within natural language understanding \citep{wang2018glue}, including intent prediction \citep{chen2019bert,castellucci2019multi}. However, recent work has demonstrated that leveraging dialog-specific pre-trained models, such as ConveRT \citep{henderson2019convert,casanueva2020efficient} or \textsc{ConvBERT} \citep{dialoglue} obtains better results. In this paper, we build on a strong pre-trained conversational encoder (\textsc{ConvBERT}) (1) by enhancing its ability to effectively capture the semantics of the input through \textbf{observers} and (2) by re-formulating the problem of intent prediction as a sentence similarity task through \textbf{example-driven training} in an effort to better leverage the strengths of language encoders and facilitate generalizability.

\subsection{Observers}

Analysis of BERT's attention weights shows that a significant amount of attention is attributed to special tokens, which have no inherent meaning \citep{clark2019does,kovaleva2019revealing}. We address this problem by disentangling BERT's attention through the use of observers. There have been several avenues of recent work that have explored disentangling the attention mechanism in Transformers. \citet{chen-etal-2019-semantically} explore disentangling the attention heads of a Transformer model conditioned on dialog acts to improve response generation. \citet{he2020deberta} disentangle the attention corresponding to the words and to the position embeddings to attain performance gains across several NLP tasks. \citet{guo2019star} propose an alternative to the fully-connected attention, wherein model complexity is reduced by replacing the attention connections with a star shaped topology.

\subsection{Example-Driven Training}

Recent efforts in NLP have shown the effectiveness of relying on an explicit set of \textit{nearest neighbors} to be effective for language modelling \citep{khandelwal2019generalization}, question answering \citep{kassner2020bert} and knowledge-grounded dialog \citep{fan2020augmenting}. However, these approaches condition on examples only during inference or in a non end-to-end manner. In contrast, we \textit{train} the encoder to classify utterances by explicitly reasoning over a set of examples. 

The core idea of example-driven training is similar to that of metric-based meta learning which has been explored in the context of image classification, wherein the objective is to learn a kernel function (which in our case is BERT) and use it to compute similarity to a support set \citep{koch2015siamese,vinyals2016matching,snell2017prototypical}. In addition to being the first to extend this approach to the task of intent prediction, the key difference of example-driven training is that we use a pre-trained language encoder \citep{dialoglue} as the underlying sentence similarity model (i.e., kernel function). \citet{ren2020intention} leverage a triplet loss for intent prediction, which ensures that their model learns similar representations for utterances with the same intent. We go beyond this, by performing end-to-end prediction in an example-driven manner. Our non-parametric approach for intent prediction allows us to attain SoTA results and facilitate generalizability to unseen intents and across datasets.

\section{Conclusion}

In order to enhance the generalizability of intent prediction models, we introduce (1) observers and (2) example-driven training. We attain SoTA results on three datasets in both full data and the few shot settings. Furthermore, our proposed approach exhibits the ability to transfer to unseen intents and across datasets without any additional training, highlighting its generalizability. We carry out a probing experiment that shows the representations produced by observers to better capture the semantic information in the input. 

There are several avenues for future work. (1) Observers and example-driven training can be extended beyond intent prediction to tasks like slot filling and dialog state tracking. (2) Since observers are disentangled from the attention graph, it is worth exploring whether it possible to force each of the observers to capture a \textit{ different} property of the input (i.e., intent, sentiment, domain, etc.). (3) Our mechanism for measuring sentence similarity in our example-driven formulation can be improved. 

\section{Ethical Considerations}

Our paper presents several approaches for improving performance on the task of intent prediction in task-oriented dialogs. We believe that neither our proposed approaches nor the resulting models have cause for ethical concerns. There is limited potential for misuse. Given the domain of our data (i.e., task-oriented dialogs), failure of the models will not result in harmful consequences. Our paper relies on significant experimentation, which may have result in a higher carbon footprint, however this is unlikely to be drastically higher than the average NLP paper. 

\bibliographystyle{acl_natbib}
\bibliography{custom}

\clearpage
\appendix

\section{Examples}

Table 6 shows examples of predictions on the \textsc{HWU} corpus using both observers and example-driven. These examples show that semantically similar example utterances are identified, particularly when using observers. Furthermore, the examples in Table 6 show that explicitly reasoning over examples makes intent classification models more interpretable.

\definecolor{green}{HTML}{15B01A}

\begin{table*}[h]
    \renewcommand*{\arraystretch}{1.2}
    \centering
    {
    \begin{tabular}{l}
    \toprule
        \textbf{Utterance:} It is too loud. Decrease the volume \\
        \textbf{Intent:} \texttt{audio-volume-down}  \\ \hline
        
        \textbf{Model:} \textsc{ConvBERT + MLM + \textit{Example}} \\ 
        \textbf{Predicted Intent:} \color{red}\texttt{audio-volume-up} \\
        \textbf{Nearest Examples:}  \\
        \hspace{7px} Make sound louder (\texttt{audio-volume-up}) \\
        \hspace{7px} Your volume is too high, please repeat that lower (\texttt{audio-volume-down}) \\
        \hspace{7px} Too loud (\texttt{audio-volume-down}) \\
        \hspace{7px} Can you speak a little louder (\texttt{audio-volume-up}) \\ \hline
        
        \textbf{Model:} \textsc{ConvBERT + MLM + \textit{Example}  + \textit{Observers}} \\ 
        \textbf{Predicted Intent:} \color{green}\texttt{audio-volume-down\textbf{}} \\
        \textbf{Nearest Examples:}  \\
        \hspace{7px} It's really loud can you please turn the music down (\texttt{audio-volume-down}) \\
        \hspace{7px} Up the volume the sound is too low (\texttt{audio-volume-up}) \\
        \hspace{7px} Too loud (\texttt{audio-volume-down}) \\
        \hspace{7px} Decrease the volume to ten (\texttt{audio-volume-down}) \\
        
        \bottomrule
        
        \textbf{Utterance:} Please tell me about the historic facts about India \\
        \textbf{Intent:} \texttt{qa-factoid}  \\ \hline
        
        \textbf{Model:} \textsc{ConvBERT + MLM + \textit{Example}} \\ 
        \textbf{Predicted Intent:} \color{red}\texttt{general-quirky} \\
        \textbf{Nearest Examples:}  \\
        \hspace{7px} How has your life been changed by me (\texttt{general-quirky}) \\
        \hspace{7px} Is country better today or ten years ago? (\texttt{general-quirky}) \\
        \hspace{7px} What happened to Charlie Chaplin? (\texttt{general-quirky}) \\
        \hspace{7px} How does production and population affect us? (\texttt{general-quirky}) \\ \hline
        
        \textbf{Model:} \textsc{ConvBERT + MLM + \textit{Example}  + \textit{Observers}} \\ 
        \textbf{Predicted Intent:} \color{green}\texttt{qa-factoid} \\
        \textbf{Nearest Examples:}  \\
        \hspace{7px} Tell me about Alexander the Great (\texttt{qa-factoid}) \\
        \hspace{7px} Give me a geographic fact about Vilnius (\texttt{qa-factoid}) \\
        \hspace{7px} Tell me about Donald Trump (\texttt{qa-factoid}) \\
        \hspace{7px} I want to know more about the upcoming commonwealth games (\texttt{qa-factoid}) \\
        \bottomrule
        
    \label{tab:examples}

    \end{tabular}
    \caption{Examples of predictions on the \textsc{HWU} corpus with both observers and example-driven training.}}
\end{table*}

\section{Ablations}

We carry out ablations over the number of observers used to train and evaluate the models. Furthermore, we vary the number of examples seen at \textit{inference time}, as a percentage of the set of training examples. The results shown in Table \ref{ablation} demonstrate that while having more observers helps, even a single observer provides benefits. This suggests that the observed performance gain (shown in Table \ref{results-table}) is primarily a consequence of the disentangled attention rather than averaging over multiple observers. 

The ablation over the number of examples used at inference time demonstrates that the models perform reasonably well with much fewer examples (e.g., 5\% is $<$1000 examples or approximately 5 per intent). The performance drop in the few-shot experiments suggests that it is important to train with more data, however the results in Table \ref{ablation} demonstrate that it not necessarily important to have all of the examples at inference time.

\begin{table*}[!t]
\centering
\renewcommand*{\arraystretch}{1.1}
{
\begin{tabularx}{\linewidth}{l X  X  X}
\toprule
  \textbf{Setting} & \multicolumn{1}{c}{\bf \textsc{banking77}} & \multicolumn{1}{c}{\bf \textsc{clinc150}} & \multicolumn{1}{c}{\bf \textsc{hwu64}} \\
  \cmidrule(lr){2-2} \cmidrule(lr){3-3} \cmidrule(lr){4-4}

 &    & &  \\[-2ex] 
 
\textsc{Observers = 20; Examples = 100\%} & \multicolumn{1}{c}{\textbf{93.83}}   & \multicolumn{1}{c}{97.31}  & \multicolumn{1}{c}{\textbf{93.03}} \\

\textsc{Observers = 10; Examples = 100\%} & \multicolumn{1}{c}{93.60}   & \multicolumn{1}{c}{\textbf{97.62}}  & \multicolumn{1}{c}{92.01} \\

\textsc{Observers = 5; Examples = 100\%} & \multicolumn{1}{c}{93.37}   & \multicolumn{1}{c}{97.38}  & \multicolumn{1}{c}{92.19} \\

\textsc{Observers = 1; Examples = 100\%} & \multicolumn{1}{c}{\textbf{93.83}}   & \multicolumn{1}{c}{97.33}  & \multicolumn{1}{c}{92.57} \\

\textsc{Observers = 20; Examples = 50\%} & \multicolumn{1}{c}{\textbf{93.83}}   & \multicolumn{1}{c}{97.31}  & \multicolumn{1}{c}{\textbf{93.03}} \\

\textsc{Observers = 20; Examples = 10\%} & \multicolumn{1}{c}{92.86}   & \multicolumn{1}{c}{97.24}  & \multicolumn{1}{c}{92.38} \\

\textsc{Observers = 20; Examples = 5\%} & \multicolumn{1}{c}{92.82}   & \multicolumn{1}{c}{96.95}  & \multicolumn{1}{c}{92.57} \\

\textsc{Observers = 20; Examples = 1\%} & \multicolumn{1}{c}{80.40}   & \multicolumn{1}{c}{68.37}  & \multicolumn{1}{c}{73.79} \\

\bottomrule
\end{tabularx}}%
\caption{Ablation over the number of observers (during both training and testing) and the number of examples (only during testing) used for the \textsc{ConvBERT + MLM + Example-Driven +  Observers} model. The percentage of examples refers to the proportion of the \textit{training set} that is used as examples for the model at evaluation time.}
\label{ablation}
\end{table*}

\end{document}